\documentclass[sigconf]{acmart}

\usepackage{array}
\usepackage{multirow}
\usepackage{xspace}
\usepackage{enumerate}
\usepackage{pifont}

\newcommand{\cmark}{\ding{51}}
\newcommand{\xmark}{\ding{55}}

\newcommand{\modify}[1]{#1}

\newcommand{\lstd}[1]{\textcolor{darkgray}{$\pm$#1}} 
\newcommand{\bwd}[1]{$\text{\textbf{#1}}^\dagger$}  
\newcommand{\mt}[1]{\textcolor[rgb]{0.6, 0, 0}{\textbf{#1}}}  
\newcommand{\mi}[1]{\textcolor[rgb]{0.8, 0.48, 0}{\textbf{#1}}}  
\newcommand{\et}[1]{\textcolor[rgb]{0, 0, 0.6}{\textbf{#1}}}  
\newcommand{\ei}[1]{\textcolor[rgb]{0, 0.4, 0}{\textbf{#1}}}  
\AtBeginDocument{%
  }

\copyrightyear{2023}
\acmYear{2023}
\setcopyright{acmlicensed}
\acmConference[MM '23]{Proceedings of the 31st ACM International Conference on Multimedia}{October 29-November 3, 2023}{Ottawa, ON, Canada}
\acmBooktitle{Proceedings of the 31st ACM International Conference on Multimedia (MM '23), October 29-November 3, 2023, Ottawa, ON, Canada}
\acmPrice{15.00}
\acmDOI{10.1145/3581783.3612575}
\acmISBN{979-8-4007-0108-5/23/10}

\begin{document}

\title{DRIN: Dynamic Relation Interactive Network for Multimodal Entity Linking}


\author{Shangyu Xing}
\authornote{Both authors contributed equally to this research.}
\email{xsy@smail.nju.edu.cn}
\author{Fei Zhao}
\authornotemark[1]
\email{zhaof@smail.nju.edu.cn}
\affiliation{%
  \institution{State Key Laboratory for Novel Software Technology, Nanjing University}
  \city{Nanjing}
  \country{China}
}


\author{Zhen Wu}
\authornote{Corresponding author.}
\email{wuz@nju.edu.cn}
\affiliation{
  \institution{State Key Laboratory for Novel Software Technology, Nanjing University}
  \city{Nanjing}
  \country{China}
}
\author{Chunhui Li}
\email{lich@smail.nju.edu.cn}
\affiliation{
  \institution{State Key Laboratory for Novel Software Technology, Nanjing University}
  \city{Nanjing}
  \country{China}
}
\author{Jianbing Zhang}
\email{zjb@nju.edu.cn}
\affiliation{
  \institution{State Key Laboratory for Novel Software Technology, Nanjing University}
  \city{Nanjing}
  \country{China}
}
\author{Xinyu Dai}
\email{daixinyu@nju.edu.cn}
\affiliation{
  \institution{State Key Laboratory for Novel Software Technology, Nanjing University}
  \city{Nanjing}
  \country{China}
}


\begin{abstract}
Multimodal Entity Linking (MEL) is a task that aims to link ambiguous mentions within multimodal contexts to referential entities in a multimodal knowledge base. 
Recent methods for MEL adopt a common framework: they first interact and fuse the text and image to obtain representations of the mention and entity respectively, and then compute the similarity between them to predict the correct entity. However, these methods still suffer from two limitations: first, as they fuse the features of text and image before matching, they cannot fully exploit the fine-grained alignment relations between the mention and entity. Second, their alignment is static, leading to low performance when dealing with complex and diverse data.
To address these issues, we propose a novel framework called Dynamic Relation Interactive Network (DRIN) for MEL tasks. DRIN explicitly models four different types of alignment between a mention and entity and builds a dynamic Graph Convolutional Network (GCN) to dynamically select the corresponding alignment relations for different input samples. 
Experiments on two datasets show that DRIN outperforms state-of-the-art methods by a large margin, demonstrating the effectiveness of our approach. \modify{Our code and datasets are publicly available\footnote{\url{https://github.com/starreeze/drin}.}}.
\end{abstract}


\maketitle

\section{Introduction}
\begin{figure}
    \centering
    \includegraphics[width=0.48\textwidth]{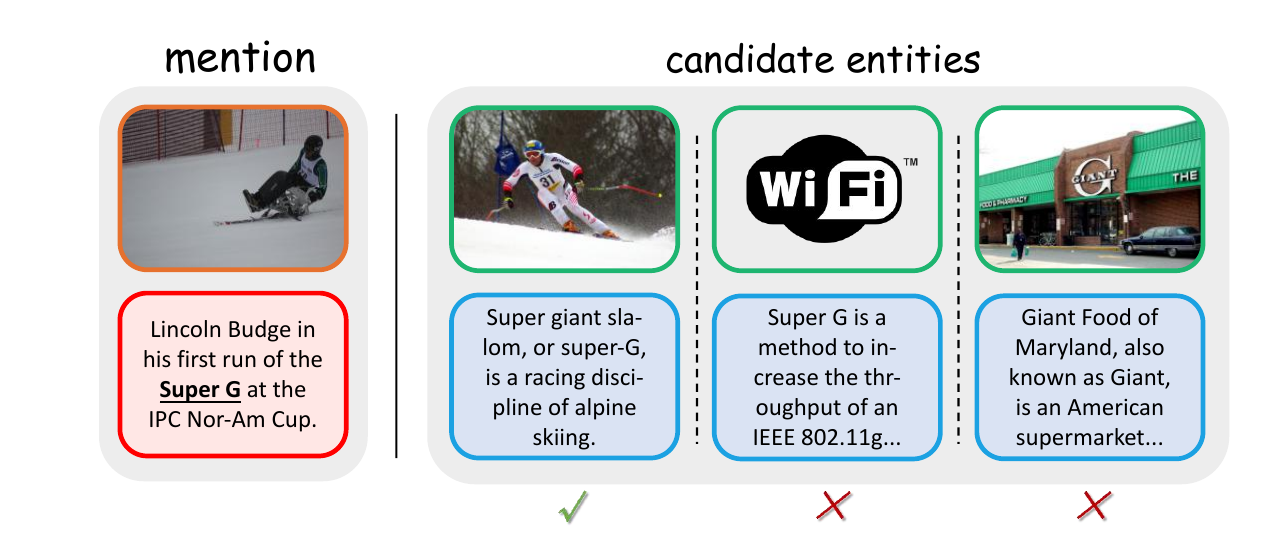}
    \caption{an example of Multimodal Entity Linking. Different colors represent different types of features: color \mt{red} for mention textual context (mention text), color \mi{orange} for mention visual context (mention image), color \et{blue} for entity textual description (entity text), and color \ei{green} for entity visual description (entity image).}
    \label{fig:task}
\end{figure}

\begin{figure}
    \centering
    \includegraphics[width=0.48\textwidth]{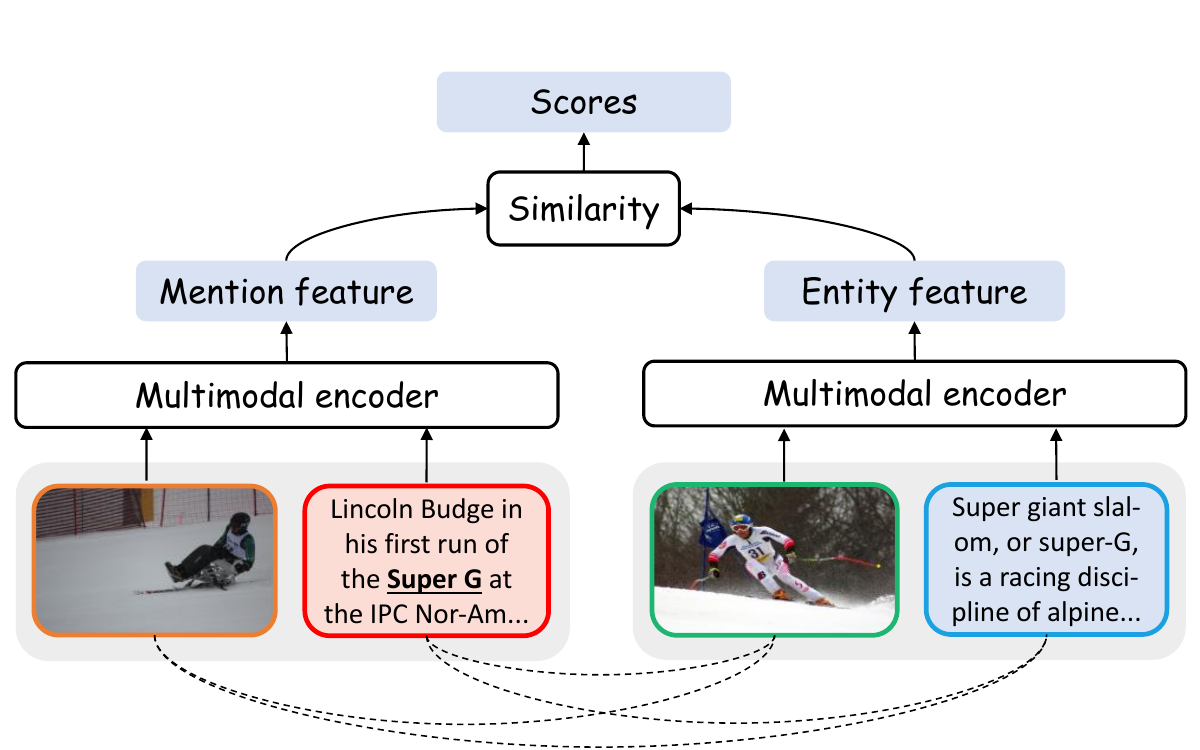}
    \caption{the common framework of previous methods, in which they implicitly align text and images of mention and entity, represented as the dashed curves.}
    \label{fig:previous}
\end{figure}

\begin{figure*}
    \centering
    \includegraphics[width=0.82\textwidth]{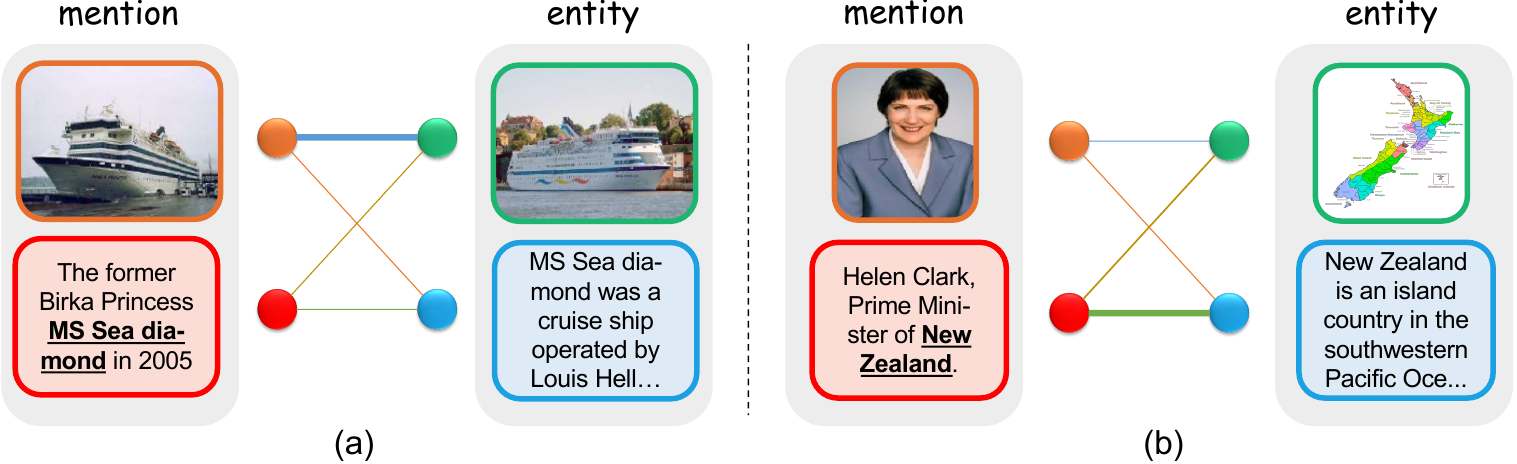}
    \caption{samples of mentions with their correct entities. The circles beside text and image represent the corresponding features, the lines indicate explicit alignments, and among them the thick ones are important alignments. Sample (a) mainly relies on image-to-image alignment, while sample (b) depends on text-to-text alignment.}
    \label{fig:samples}
\end{figure*}

Multimodal Entity Linking (MEL) is an important research area in natural language processing, with the goal of linking ambiguous mentions in multimodal contexts to entities in a multimodal knowledge base \cite{DBLP:conf/mm/GanLWWHH21}. \modify{As a basic task of multimodal information retrieval, MEL has a wide range of real-world applications, including multimodal dialog system and visual question answering \cite{DBLP:conf/nips/LuYBP16}.} The biggest challenge of MEL is the ambiguity of the mention, i.e., a word or phrase may convey different meanings under different circumstances, so in MEL task visual context is needed to help disambiguate it. For instance, as illustrated in Figure \ref{fig:task}, the mention phrase ``Super G'' can be interpreted as many entities, such as skiing, WLAN protocol, or food market. It is difficult to find the correct entity using the text modality alone. However, with its image considered, we can understand that this ``Super G'' is related to skiing. In this way, disambiguation is much easier.

So far, most existing methods divide MEL into two stages: candidate retrieval and entity disambiguation \cite{DBLP:conf/dasfaa/ZhangLY21,DBLP:conf/mm/GanLWWHH21,DBLP:conf/acl/WangTGLWYCX22,DBLP:conf/sigir/WangWC22}. In the first stage, they roughly calculate the similarity between the mention and all the entities in the knowledge base with basic approaches (e.g., edit distance \cite{DBLP:conf/sigir/WangWC22,DBLP:conf/dasfaa/ZhangLY21}, simple encoders \cite{DBLP:conf/acl/WangTGLWYCX22} or statistical methods \cite{DBLP:conf/mm/GanLWWHH21,DBLP:conf/acl/WangTGLWYCX22}) to retrieve the Top-K candidate entities that are most similar to the mention. In the second stage, detailed multimodal information is used to predict the correct entity from the candidate set constructed before \cite{DBLP:conf/dasfaa/ZhangLY21,DBLP:conf/mm/GanLWWHH21,DBLP:conf/acl/WangTGLWYCX22,DBLP:conf/sigir/WangWC22}. Recent methods for this stage all adopt a common framework (shown in Figure \ref{fig:previous}): they first interact and fuse the text and image of the mention to obtain the mention representation, and then calculate the entity representation with its text and image in a similar way. Finally, they compute the similarity between them to make prediction.

Although achieving good results, their ``fuse and then match'' mechanism implicitly models the alignment relations between the <\mt{text}, \mi{image}> of mention and the <\et{text}, \ei{image}> of candidate entity, which actually includes four types of different alignments, i.e., mention \mt{text} and entity \et{text}, mention \mt{text} and entity \ei{image}, mention \mi{image} and entity \et{text}, mention \mi{image} and entity \ei{image}. This brings two potential drawbacks:

Firstly, it is difficult for the model to model the fine-grained relations of the mention and entity. As the features of text and images are fused before matching, some fine-grained features are mixed and weakened, so they cannot be easily aligned between the mention and entity. For example, as shown in Figure \ref{fig:samples}(a), the previous implicit alignment method fuses features of the ``ship'' in the mention image with mention text features. Thus, the fused features cannot be aligned with the ship in the entity image. However, if the two images are explicitly associated, it is easy to find the fact that ``the main visual objects in the images are both ships''. This clue is crucial for MEL task because it indicates that the mention and entity refer to the same object. Therefore, a high-performing model need to explicitly model the alignment relations between mention and entity.

Secondly, their alignment is static, which results in low performance when dealing with complex and diverse data, as different samples often rely on different types of alignment. For instance, some depend on \mt{text}-to-\et{text} alignment, while others mainly rely on \mi{image}-to-\ei{image} alignment. As shown in Figure \ref{fig:samples}(a), the text contains little useful information, and this sample mainly relies on the alignment relation of images, i.e., discovering that the visual objects are both ships. Conversely, the image in Figure \ref{fig:samples}(b) does not contain adequate information to indicate that ``New Zealand'' is a country (rather than a sports team); only by focusing on the ``prime minister'' in the text can it be associated with the country. So an effective model should be able to adaptively select the corresponding alignment based on different input samples.

To address these issues, we propose the Dynamic Relation Interactive Network (DRIN). For the first issue, we \emph{explicitly} model four different types of alignment, which enables DRIN to learn fine-grained alignment relations between mention and entity. For the second issue, we build a \emph{dynamic} GCN, which improves the model's ability to handle varied data. Concretely,  We treat the text and image in mention, as well as the text and images in candidate entities, as vertices, and the four different types of alignment relations as edges. By iteratively updating vertex features and edge weights, we can dynamically select the corresponding relations for different input samples. Experiments on two datasets show that DRIN outperforms state-of-the-art methods by a large margin, demonstrating the effectiveness of our method.

Overall, our contribution can be summarized as follows:
\begin{enumerate}[1)]
    \item We are among the first to adopt dynamic explicit fine-grained alignments to the MEL task, which improves the performance when dealing with complex and diverse data;
    \item We propose a novel dynamic relation interaction framework that updates features and relations dynamically on a GCN, resulting in more accurate and robust representations;
    \item Experiments on two public datasets demonstrate that DRIN outperforms previous state-of-the-art works, and further analysis verifies the validity of our proposed network.
\end{enumerate}

\section{Related Work}
\subsection{Entity Linking}
Recent methods for Entity Linking (EL) all employ neural networks. They first use text encoders to obtain context-aware representations of mention and entity, and then calculate similarities between them to further obtain the final probabilities. For text encoders, they usually use LSTM \cite{DBLP:journals/neco/HochreiterS97} or BERT \cite{DBLP:journals/corr/abs-1810-04805}. Similarity measures include dot product \cite{DBLP:conf/emnlp/GaneaH17,DBLP:conf/emnlp/GuptaSR17,DBLP:conf/conll/KolitsasGH18,DBLP:conf/emnlp/WuPJRZ20} and cosine similarity \cite{DBLP:conf/naacl/Francis-LandauD16,DBLP:conf/conll/GillickKLPBIG19,DBLP:conf/ijcai/SunLTYJW15}. When calculating final probabilities, some methods apply an additional feed-forward network layer and a softmax layer \cite{DBLP:conf/naacl/Francis-LandauD16,DBLP:conf/emnlp/GaneaH17,DBLP:journals/corr/abs-1908-05762}.

However, these methods are designed to deal with text only, and cannot handle multimodal tasks. As multimodal data becomes more and more important recently, there is a growing need for new methods that can handle Multimodal Entity Linking.

\subsection{Multimodal Entity Linking}
\label{sec:mel}
Multimodal Entity Linking (MEL) is an extension of EL that utilizes additional multimodal information (e.g., images, audios or videos) to help disambiguate entities. Currently, most studies focus on tasks where only text and images are involved.

Based on previous works, the task of MEL can be separated into two categories. The difference between them is that their images play different roles: the first category aims to link noun phrases in the mention sentence, with images as auxiliary information \cite{DBLP:conf/acl/CarvalhoMN18,DBLP:conf/dasfaa/ZhangLY21,DBLP:conf/acl/WangTGLWYCX22,DBLP:conf/sigir/WangWC22,DBLP:conf/coling/0002HLL18,DBLP:conf/emnlp/ZhangH22,DBLP:journals/corr/abs-2307-09721}, while second category links both noun phrases in the sentence and visual objects in the corresponding image respectively to the text and image in the knowledge base \cite{DBLP:conf/mm/GanLWWHH21}. Since the first category is dominant, we adopt it as our task format.

To tackle this task, Moon et al. \cite{DBLP:conf/acl/CarvalhoMN18} use a cross-modal attention mechanism to fuse features at the character, word, and image levels, before calculating similarities. Adjali et al. \cite{DBLP:conf/ecir/AdjaliBFBG20} construct a more challenging dataset on Twitter and design corresponding inter-modal interactions and loss functions. Zhang et al. \cite{DBLP:conf/dasfaa/ZhangLY21} design a two-stage mechanism to reduce the negative impact of noisy images. They calculate the relation between images and text, allowing only related images to enter subsequent steps. \modify{Zhang et al. \cite{DBLP:conf/emnlp/ZhangH22} proposed to utilize history context on social media and designed a co-attention scheme to aid the disambiguation process.} Wang et al. \cite{DBLP:conf/acl/WangTGLWYCX22} combine feature representation and statistical probability, using inter- and intra-modal attention to better fuse multimodal information. Wang et al. \cite{DBLP:conf/sigir/WangWC22} propose a gate fusion method to control the weights of different modalities, and use contrastive learning to obtain more meaningful multimodal representations.

All these works adopt a common framework: they first fuse the text and image on both the mention and entity side, and then use various techniques to match their information. This means that they implicitly model the alignment relation between the text and image of the mention and entity. Compared to this implicit alignment approach, our proposed dynamic explicit alignment approach has a superior performance in discovering fine-grained relations and handling variable data.

\subsection{Graph Convolutional Network}
The idea of graph convolutional networks (GCNs) originated from traditional convolutional neural networks, which extended the convolution operation to graph structures. Traditional GCNs \cite{DBLP:journals/corr/KipfW16} use the adjacency matrix to convolve the information of neighboring vertices onto the current vertex, After several iterations, it can perceive the graph structure with surrounding information. On this basis, improved structures are proposed. GAT \cite{DBLP:journals/corr/abs-1710-10903} only accepts vertex features as input and uses attention mechanisms between pairs of vertices to replace traditional edges. KE-GCN \cite{DBLP:conf/www/YuYZW21} embeds representations of both vertices and edges as vectors, and they are iteratively updated during convolution.

Cao et al. \cite{DBLP:conf/coling/0002HLL18} first introduced GCNs into the EL task. They construct an entity graph connecting mention context to candidate entities, and then disambiguate with the help of contextual information. However, their model only uses textual relations and cannot handle multimodal tasks. To the best of our knowledge, we are the first to apply graph convolutional networks to MEL tasks.

\section{Methodology}
\begin{figure*}
    \centering
    \includegraphics[width=\textwidth]{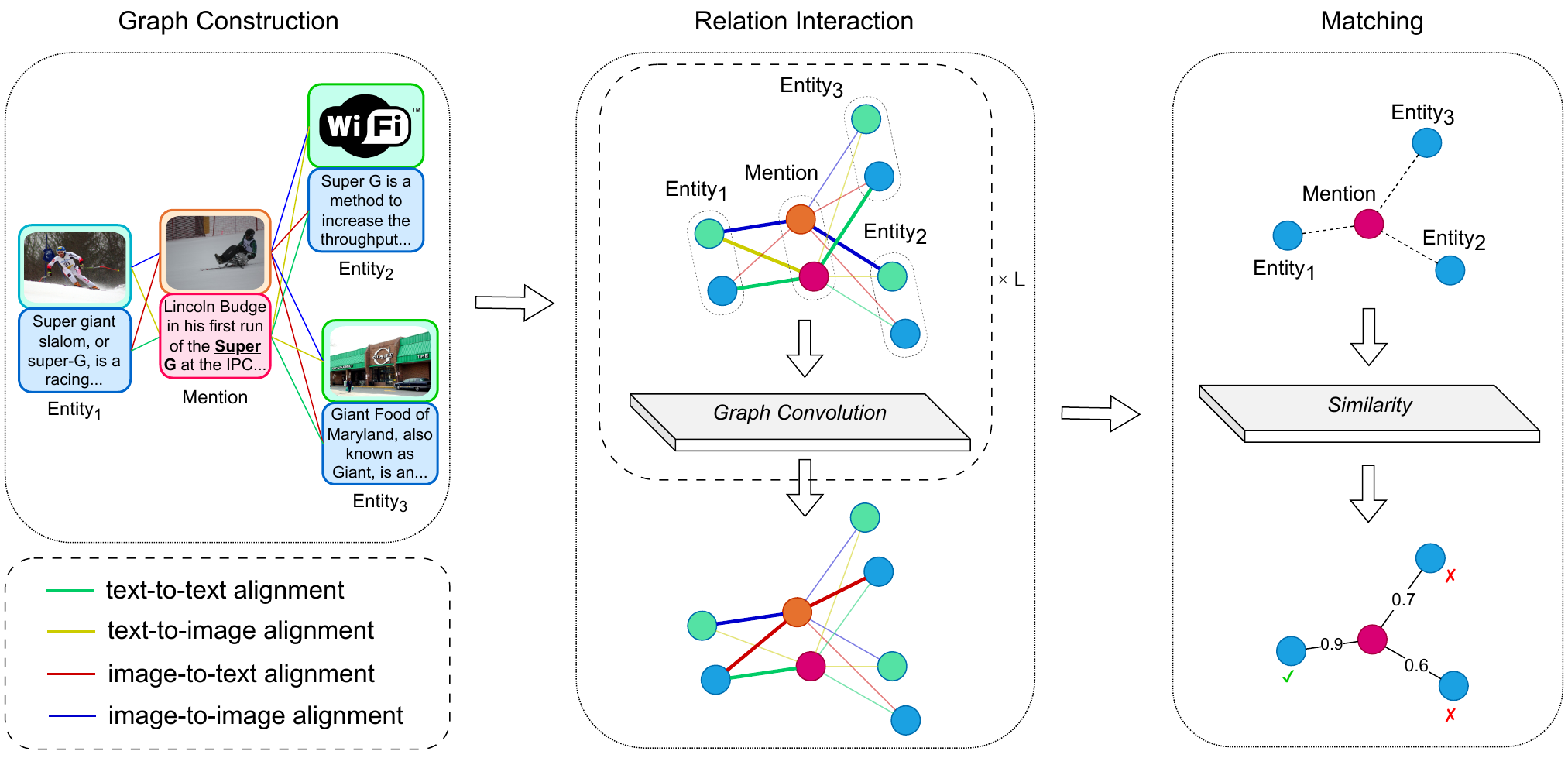}
    \caption{an overview of the DRIN model with candidate set size r=3. First, we extract text and image features and calculate similarities between them to construct an undirected graph. Second, we perform dynamic graph convolutions on it to iteratively update the vertex representations and edge weights. Third, we extract the text vertex as their final representation, and calculate the similarity of them to make predictions.}
    \label{fig:method}
\end{figure*}

\subsection{Problem Formulation}
The task of Multimodal Entity Linking is to map a mention to its corresponding entity in a knowledge base. To simplify the problem, we assume each mention or entity contains only one image, following the previous works \cite{DBLP:conf/acl/WangTGLWYCX22,DBLP:conf/sigir/WangWC22}. 

As mentioned earlier, MEL is generally a two-stage task, with the second stage being harder than the first. For the first stage, we adopt the same method as previous works \cite{DBLP:conf/sigir/WangWC22,DBLP:conf/acl/WangTGLWYCX22} to extract candidate entities. We will describe in detail how we construct the candidate set and make comparison in Section \ref{sec:compare}. The second stage is our focus, which involves linking a mention to its entity from a candidate set constructed in the first stage for each mention.

Formally, given a mention $m$, let $C(m)$ represent its candidate entity set, which usually contains tens of entities. Let $r$ denotes the size of the candidate set, i.e., $r=|C(m)|$. Mention $m$ is characterized by its context $T_m$(mention textual context, i.e., mention text) and $V_m$(mention visual context, i.e., mention image). An entity $e\in C(m)$ is characterized by its description $T_e$(entity textual description, i.e., entity text) and $V_e$(entity visual description,  i.e., entity image). The referent entity of mention $m$ is predicted through:
\begin{equation}
    e^*(m)=\mathop{\arg\max}\limits_{e\in C(m)}\text{sim}(T_m,V_m;T_e,V_e),
\end{equation}
where $\text{sim}(\cdot)$ is the similarity between the mention and entity.

\subsection{Overview}
As illustrated in Figure \ref{fig:method}, DRIN consists of three modules: graph construction, relation interaction, and matching.

In the first module, we extract visual features using ResNet and embed textual features with BERT \cite{DBLP:journals/corr/abs-1810-04805} for both mentions and entities. These four types of features serve as the initial values of the corresponding vertices in GCN. After that, we define four types of alignment relations between mentions and entities, namely text-to-text, text-to-image, image-to-text, and image-to-image, and calculate their similarities as the initial edge weights in GCN.

In the second module, we perform relation interaction on a dynamic GCN. A graph is constructed on a mention and its candidate entities, where vertices represent the text and images of the mention and entities, and edges represent the relations between them. After obtaining the initial values in the first step, we perform dynamic convolution on the constructed graph to iteratively update vertex features and edge weights.

In the last module, we retrieve the text vertex features of mentions and entities from the GCN, which are already aware of multimodal contextual information after a few iterations, and calculate the cosine similarity scores between them to obtain the final linking probabilities of candidate entities. The entity with the largest probability is predicted as the linking target of the mention.

\subsection{Graph Construction}
\subsubsection{Vertex Feature}
\label{sec:vertex}
We first extract textual and visual features to initialize the vertices, including mention text, mention image, entity text and entity image. 

As a strong text encoder, pre-trained model BERT\cite{DBLP:journals/corr/abs-1810-04805} is widely used in different NLP tasks. For mention sentence, we use BERT encoder to obtain its context-aware representations and extract mention phrase token features from it. After that, they are average pooled into a vector and linearly mapped to a subspace. Formally,
\begin{equation}
    V_{mt}=W_{mt}\frac{1}{e-b}\sum_{i=b}^{e-1}\left(\text{BERT}(T_m)[i]\right),
\end{equation}
where $W_{mt}$ is a trainable weight, $b,e$ denote the beginning and ending position of the mention phrase in the sentence, and $V_{mt}$ is the representation for the mention text.

For entity description, we also apply the BERT encoder to obtain its feature. Next, we extract the feature of the first token as its representation and map it to the subspace:
\begin{equation}
    V_{et}=W_{et}\text{BERT}(T_e)_{[CLS]},
\end{equation}
where $V_{et}$ is the representation for entity text.

In addition to textual features, we also encode visual features. Most of the previous methods in the MEL task use the pre-trained ResNet \cite{DBLP:journals/corr/HeZRS15} model to extract image features. For a fair comparison, we use the same image encoder as previous models. Specifically, given an image, we first resize it to 224×224 pixels, and then use the image recognition model ResNet to obtain the pooler output of the last convolutional layer. After that, the feature is also mapped to the subspace. Formally, we obtain the visual features through
\begin{equation}
    V_{mv}=W_{mv}\text{ResNet}(V_m),
\end{equation}
\begin{equation}
    V_{ev}=W_{ev}\text{ResNet}(V_e),
\end{equation}
where $W_{mv},W_{ev}$ are trainable weights, ResNet$(\cdot)$ is the ResNet image encoder whose output is a pooled vector, and $V_{mv},V_{ev}$ is vertex features for mention image and entity image, respectively.

\subsubsection{Edge Relation}
\label{sec:edge}
To leverage fine-grained alignment between mention and entity, we explicitly model four types of relations: mention text and entity text, mention text and entity image, mention image and entity text, mention image and entity image. We use the these relations to build the corresponding edges in GCN.

\textbf{\mt{Text}-to-\et{text} alignment}: the semantic meaning of mention is implied by its context\cite{DBLP:conf/coling/0002HLL18}. If mention textual context and entity textual description is similar, it is likely that they refer to the same object. Therefore, by connecting them with an edge in GCN, they can aggregate information from each other and learn better representations. We model the text-to-text relation $R_{tt}\in \mathbb{R}$ with the similarity of the BERT feature of their first token, as:
\begin{equation}
    R_{tt}=\cos(\text{BERT}(T_m)_{[CLS]},\text{BERT}(T_e)_{[CLS]}).
\end{equation}

\textbf{\mt{Text}-to-\ei{image} alignment}: both the mention textual context and entity image imply the topic of the information\cite{DBLP:conf/coling/0002HLL18}. If a mention and a candidate entity are under the same topic, it is more probable that they refer to the same object. For example, as shown in Figure \ref{fig:task}, the words ``first run'' and ``cup'' indicate that the topic of this mention is sports. We can also learn from the first entity image, in which a man is skiing, that this entity is also in the sports topic. Therefore, they two can use each other's information to enhance their representations. To capture this type of alignment relations, we use the CLIP\cite{DBLP:journals/corr/abs-2103-00020} model to obtain the text-image correlation $R_{tv}\in \mathbb{R}$, as it is a powerful pre-trained multimodal model that can be used to calculate text and image similarity:
\begin{equation}
    R_{tv}=\text{CLIP}(T_m, V_e).
\end{equation}

\textbf{\mi{Image}-to-\et{text} alignment}: the mention usually appears in its image, so the corresponding visual object can represent the semantic meaning of the mention. If that is similar with the entity description, there is a greater chance that they refer to the same object. We also use CLIP to get $R_{vt}\in \mathbb{R}$:
\begin{equation}
    R_{vt}=\text{CLIP}(V_m, T_e).
\end{equation}

\textbf{\mi{Image}-to-\ei{image} alignment}: if the mention image and the entity image contain the same visual object, chances are that this object is what they both refer to. \modify{Inspired by \cite{DBLP:conf/mm/ZhaoLWXD22}}, we first apply the pre-trained object detection model faster-RCNN \cite{DBLP:conf/nips/RenHGS15} to extract the top-k visual object regions of both images with the scores of confidence, denoted as 
\begin{equation}
    V_m^{obj}=\{v_m^i\}_{i=1}^k,~~V_m^{score}=\{s_m^i\}_{i=1}^k,
\end{equation}
and
\begin{equation}
    V_e^{obj}=\{v_e^i\}_{i=1}^k,~~V_e^{score}=\{s_e^i\}_{i=1}^k,
\end{equation}
where $k$ is the number of visual objects extracted for both mention and entity. 
Then we calculate the weighted average of the similarities between the two sets of objects to obtain image-to-image relation $R_{vv}\in \mathbb{R}$:
\begin{equation}
    R_{vv}=\frac{\sum_{i=1}^k \sum_{j=1}^k s_m^i s_e^j \cos(v_m^i,v_e^j)}{\sum_{i=1}^k s_m^i \sum_{i=1}^k s_e^i},
\end{equation}
where $\cos(\cdot)$ is the cosine similarity function.

\subsection{Relation Interaction}
Relation interaction is designed to enhance the multimodal representation of a vertex with information from its neighboring vertices.
We construct our relation interaction module based on a dynamic GCN. Formally, it is defined on an undirected graph $G = (V, E)$, where $V(|V|=n)$ is a set of vertices representing text and images of a mention and all its candidate entities, and $E$ is a set of edges specified by the four types of relations between them. Let $H\in \mathbb{R}^{n\times d}$ be a matrix containing the features of all $n$ vertices, where $d$ is the dimension of the feature vectors, and row $H_i\in \mathbb{R}^d$ is the feature vector of the i-th vertex initialized in Section \ref{sec:vertex}. Let $A\in \mathbb{R}^{n\times n}$ be the adjacent matrix that is initialized via the edge relations in Section \ref{sec:edge}. A GCN layer is a nonlinear transformation that maps from ($H^l,A^l$) to ($H^{l+1},A^{l+1}$) , defined as:
\begin{equation}
    H_i^{l+1}=\sigma\left(\sum_{j=1}^n A_{ij}^lW_h^lH_j^l\right)+H_i^l,
\end{equation}
\begin{equation}
    A_{ij}^{l+1}=M*\left(\sigma\left((W_a^lH_i^{l+1})^T(W_a^lH_j^{l+1})\right)+A_{ij}^l\right),
\end{equation}
where $l$ is the current layer index, $W_h^l,W_a^l$ are trainable parameters, $\sigma$ is a non-linear activation function, * is the element-wise multiplication, and $M\in \{0,1\}^{n\times n}$ is a mask matrix defined as
\begin{align*}
    M_{ij}=\begin{cases}
        1, \text{if relation }(i,j)\text{ is one of the four types specified,}\\
        0, \text{otherwise.}
    \end{cases}
\end{align*}

After obtaining multimodal features and similarities in the first step, we use them to initialize vertex representations $H^0$ and edge weights $A^0$. Then, we feed them into the dynamic GCN with $L$ layers to perform relation interaction.

\subsection{Matching}
Finally, we extract text vertices of both the mention and its entities from the graph, which are now aware of multimodal context, and calculate their similarities. Formally, we denote the final mention text vertex feature as $T_m^*$, and entity text vertex features as $T_{e,i}^* (i=1,2,\dots,r)$. Then the similarity is calculated as
\begin{equation}
    S(m, e_i(m))=\cos(T_m^*, T_{e,i}^*),
\end{equation}
where $S(m, e_i(m))$ is the similarity score between the mention $m$ and its $i$-th entity. Thus the index of the predicted entity
\begin{equation}
    i^*=\mathop{\arg\max}\limits_{i\in\{1,2,\dots,r\}}S(m, e_i(m)).
\end{equation}
As a result, the finally predicted entity
\begin{equation}
    e^*(m)=e_{i^*}(m).
\end{equation}

\subsection{Loss Function}
We use margin ranking loss as the loss function. The goal of training is to maximize the similarity between the mention and its correct entity while minimize that of other entities. Formally, the loss is specified by
\begin{equation}
    \mathcal{L}=\sum\max(S_--S_++\lambda, 0),
\end{equation}
where $\lambda$ is the margin, $S_+$ is the similarity between the mention and its correct entity, and $S_-$ is the mean similarity of the mention with all entities except for the correct one in a mini-batch.

\section{Experiments}
\begin{table}[]
    \centering
    \caption{statistics of WikiMEL and WikiDiverse.}
    \begin{tabular}{lcc}
        \toprule
        & \textbf{WikiMEL} & \textbf{WikiDiverse} \\
        \midrule
        \# Num. of samples & 22.1k & 9.8k \\
        \# Num. of menitons & 26.6k & 19.5k \\
        \# Average text length & 8.2 & 10.1 \\
        \# Average mentions per sample & 1.2 & 2.0 \\
        \# Num. of candidates per mention & 100 & 10 \\
        \bottomrule
    \end{tabular}
    \label{tab:stat}
\end{table}

\begin{table*}[]
    \centering
    \caption{performance comparison of different methods on the WikiMEL and WikiDiverse datasets (\%). The results of DRIN are averaged on 5 runs, with the corresponding standard deviation beside. Best results are in bold. The marker $\dagger$ indicates that the significance test p-value is less then 0.05 compared with GHMFC. We cannot get some results on WikiDiverse because these models do not have their code opened and no paper reports their performance on WikiDiverse.}
    \scalebox{0.98}{
    \begin{tabular}{lcccccccc}
        \toprule
        \multirow{2}*{Model} & \multicolumn{4}{c}{WikiMEL} && \multicolumn{3}{c}{WikiDiverse} \\
        \cmidrule{2-5} \cmidrule{7-9}
        & Top-1 & Top-5 & Top-10 & Top-20 && Top-1 & Top-3 & Top-5 \\
        \midrule
        BERT   & 31.7 & 48.8 & 57.8 & 70.3 && 45.5 & 75.7 & 89.0 \\
        JMEL   & 31.3 & 49.4 & 57.9 & 64.8 && N/A  & N/A  & N/A  \\
        DZMNED & 34.7 & 53.9 & 58.1 & 70.1 && N/A  & N/A  & N/A  \\
        MEL-HI & 38.6 & 55.1 & 65.2 & 75.7 && 45.7 & 76.5 & 88.6 \\
        GHMFC  & 43.6 & 64.0 & 74.4 & 85.8 && 46.0 & 77.5 & 88.9 \\
        DRIN   & \bwd{65.5}\lstd{0.81} & \bwd{91.3}\lstd{0.52} & \bwd{95.8}\lstd{0.22} & \bwd{97.7}\lstd{0.14} && \bwd{51.1}\lstd{0.74} & \textbf{77.9}\lstd{0.65} & \textbf{89.3}\lstd{0.83} \\
        \bottomrule
    \end{tabular}}
    \label{tab:result}
\end{table*}

\subsection{Datasets}
In this part, we first review on datasets proposed by previous works, and then describe and explain our choice.

Moon et al.\cite{DBLP:conf/acl/CarvalhoMN18} proposed the first MEL dataset SnapCaptionsKB, which is composed of 12K user-generated image and textual caption pairs from social media. Adjali et al. \cite{DBLP:conf/ecir/AdjaliBFBG20} constructed their dataset Twitter-MEL by collecting Twitter posts with text and images. Zhang et al. \cite{DBLP:conf/dasfaa/ZhangLY21} collected their text-image data from Weibo and also constructed their new dataset. However, none of the above three datasets are opened, so they are not available to us.

Gan et al. \cite{DBLP:conf/mm/GanLWWHH21} proposed an open dataset M3EL, by obtaining movie reviews from IMDb and The Movie Database. However, as stated in Section \ref{sec:mel}, its task format is different from ours. Besides noun phrases in sentences, it also regards visual objects in images as mentions to be linked. As a result, we cannot use their data.

Later, Wang et al. \cite{DBLP:conf/sigir/WangWC22} proposed three new open datasets: Wiki-MEL, Richpedia-MEL and Twitter-MEL, but only WikiMEL contains at least one image for both mention and entity. Another open dataset WikiDiverse \cite{DBLP:conf/acl/WangTGLWYCX22} was proposed at the same time.

We evaluate our model on two MEL datasets: WikiMEL and WikiDiverse, as they are the only open MEL datasets that contain both mention images and entity images, as far as we know. The statistics of WikiMEL and WikiDiverse are described in Table \ref{tab:stat}.

\subsection{Compared Methods}
\label{sec:compare}
As described before, the task of MEL contains two stages, and our focus is on the second stage. For a fair comparison, we adopt the same method as previous works to extract candidate entities in the first stage. Concretely, we follow \cite{DBLP:conf/sigir/WangWC22} to use fuzzy search to extract Top-100 candidates in WikiMEL dataset, and follow \cite{DBLP:conf/acl/WangTGLWYCX22} to adopt a combined method of statistics and word features to extract Top-10 candidates in WikiDiverse dataset.

Afterwards, we apply different methods to predict the correct entity among the previously extracted candidate set. We compare our method the following unimodal and multimodal model: 

\textbf{BERT}\cite{DBLP:journals/corr/abs-1810-04805}: a unimodal method that use the pretrained model BERT to encode the mention text and entity description, and then calculate their similarity to make a prediction.

\textbf{JMEL}\cite{DBLP:conf/ecir/AdjaliBFBG20}: a multimodal method that uses fully connected layers to project the visual and textual features into an implicit joint space. They also use contrastive learning to enhance the representations.

\textbf{DZMNED}\cite{DBLP:conf/acl/CarvalhoMN18}: a multimodal method that utilizes a multimodal attention mechanism to fuse visual, textual and character level features of mention, and then use both combined and character level features to match entity representations.

\textbf{MEL-HI}\cite{DBLP:conf/dasfaa/ZhangLY21}: a multimodal method that adopts a two-stage mechanism. It first calculates the similarity of image and text, and only allows related images to enter the multimodal fusion step.

\textbf{GHMFC}\cite{DBLP:conf/sigir/WangWC22}: a multimodal method that applies contrastive learning and a fusion gate to control the weights of different modalities.

\subsection{Evaluation Metrics}
We use the top-K accuracy metric for evaluation. Given the similarities between a mention and its candidate entities, we rank the candidates based on it. If the correct entity is ranked among the top-K candidates, the sample is considered correct. The top-k accuracy is calculated as the ratio of the number of correct samples to the total number of samples. Formally, for a dataset $D$, top-K accuracy is defined as
$$
\text{TopKAcc}(k)=\frac{1}{|D|}\sum_{m\in D}\left[\sum_{i=1}^r\left[S\left(m,e^*(m)\right)<S\left(m,e_i(m)\right)\right]<k\right],
$$
where $[\cdot]$ is the Iverson bracket which evaluates to 1 if the condition inside it is true and 0 otherwise.

\subsection{Implementation Detail}
Our proposed method is implemented using the PyTorch framework \cite{DBLP:conf/nips/PaszkeGMLBCKLGA19} and trained on an NVIDIA GeForce RTX 3090Ti GPU \cite{cuda}. We use bert-base-cased \cite{DBLP:journals/corr/abs-1810-04805}, resnet-152-imagenet \cite{DBLP:journals/corr/HeZRS15} and clip-vit-base-patch32 \cite{DBLP:journals/corr/abs-2103-00020} as our encoder. The Adam optimizer \cite{DBLP:journals/corr/KingmaB14} is utilized for training, with a fixed number of epochs set to 30. We set GCN hidden dimension size as 768, number of GCN layers as 2, batch size as 64, learning rate as 0.001, and loss margin as 0.25. We report the results averaged on 5 runs along with standard deviation and tests of significance on random initialization under the aforementioned settings.

\section{Results And Discussion}
\begin{table*}[htbp]
    \centering
    \caption{performance of ablation studies on main components of DRIN (\%).}
    \scalebox{0.98}{
    \begin{tabular}{lcccccccc}
        \toprule
        \multirow{2}*{Model} & \multicolumn{4}{c}{WikiMEL} && \multicolumn{3}{c}{WikiDiverse} \\
        \cmidrule{2-5} \cmidrule{7-9}
        & Top-1 & Top-5 & Top-10 & Top-20 && Top-1 & Top-3 & Top-5 \\
        \midrule
        DRIN                           & 65.5 & 91.3 & 95.8 & 97.7 && 51.1 & 77.9 & 89.3 \\
        DRIN (w/o image-to-image edge) & 65.2 & 91.4 & 95.8 & 97.9 && 49.1 & 76.8 & 89.1 \\
        DRIN (w/o image-to-text edge)  & 64.0 & 90.1 & 95.4 & 97.9 && 49.1 & 77.1 & 87.7 \\
        DRIN (w/o text-to-image edge)  & 64.4 & 90.3 & 95.5 & 98.0 && 49.3 & 77.4 & 87.5 \\
        DRIN (w/o text-to-text edge)   & 61.2 & 88.8 & 94.4 & 97.8 && 48.6 & 75.3 & 87.3 \\
        DRIN (static edge)             & 57.8 & 86.1 & 92.3 & 95.5 && 49.0 & 77.4 & 88.6 \\
        \bottomrule
    \end{tabular}}
    \label{tab:ablation}
\end{table*}

\subsection{Main Results}
Table \ref{tab:result} presents the results of our proposed DRIN model in comparison with previous methods on the WikiMEL and WikiDiverse datasets. Since in WikiMEL a candidate set contains 100 entities and in WikiDiverse 10 entities, we report Top-1, Top-5, Top-10, Top-20 of WikiMEL, and Top-1, Top-3, Top-5 of WikiDiverse. Based on these results, we can make a couple of observations:

First, it is notable that the unimodal baseline BERT displays a relatively commendable performance. This can be attributed to its capability of obtaining context-aware representations of both mentions and entities. In certain instances, the correct entity can be identified through the utilization of textual modality alone.

Second, all multimodal methods outperform the unimodal BERT on both datasets, indicating that the visual information is useful in supplementing the textual information for the MEL task. Among the multimodal methods, GHMFC achieves the best performance, possibly due to its self-modal and cross-modal multi-head attention, which helps to learn more robust representations.

Third, our proposed DRIN model outperforms previous methods by a significant margin on both datasets. Specifically, DRIN outperforms the state-of-the-art GHMFC model by 22.4\% and 5.1\% on the Top-1 score for WikiMEL and WikiDiverse, respectively. Our model also achieves better results on other metrics. These results further reveal the effectiveness of our model.

Fourth, all methods, including ours, perform worse on the WikiDiverse dataset. This is because the mentions and images in WikiDiverse are more diverse and varied, covering a wide range of topics from locations to famous events. This diversity makes the task more challenging compared to WikiMEL, where most mentions refer to people and images are usually their photos.

\begin{table*}[htbp]
    \centering
    \caption{performance of DRIN with different number of layers (\%).}
    \scalebox{0.97}{
    \begin{tabular}{lccccccccccccccc}
        \toprule
        \multirow{2}{0.35cm}{$L$} & \multicolumn{3}{c}{WikiMEL (valid)} && \multicolumn{3}{c}{WikiMEL (test)} && \multicolumn{3}{c}{WikiDiverse (valid)} && \multicolumn{3}{c}{WikiDiverse (test)} \\
        \cmidrule{2-4} \cmidrule{6-8} \cmidrule{10-12} \cmidrule{14-16} 
        & Top-1 & Top-5 & Top-10 && Top-1 & Top-5 & Top-10 && Top-1 & Top-3 & Top-5 && Top-1 & Top-3 & Top-5 \\
        \midrule
        1 & 64.7 & 90.0 & 95.3 && 64.0 & 90.0 & 95.3 && 47.8 & 76.4 & 87.6 && 48.8 & 78.0 & 89.1 \\
        2 & 65.5 & 91.3 & 95.8 && 65.5 & 91.3 & 95.8 && 48.5 & 77.3 & 87.8 && 51.1 & 77.9 & 89.3 \\
        3 & 65.8 & 90.7 & 95.4 && 64.1 & 90.3 & 95.5 && 45.5 & 75.1 & 86.8 && 47.2 & 74.5 & 87.2 \\
        4 & 65.5 & 90.6 & 95.5 && 63.1 & 90.4 & 95.5 && 45.9 & 75.7 & 87.9 && 47.2 & 76.4 & 86.6 \\
        5 & 63.6 & 89.3 & 94.6 && 62.9 & 89.6 & 94.5 && 43.9 & 72.3 & 86.2 && 45.4 & 75.9 & 86.7 \\
        \bottomrule
    \end{tabular}}
    \label{tab:hyper}
\end{table*}

\subsection{Ablation Study}
To investigate the contributions of different modules of the model, we conduct ablation studies on two main components of DRIN. The results are shown in Table \ref{tab:ablation}, where ``w/o'' indicates the removal of the graph edges, and ``static edge''refers to fixing the edge weight at the initial value throughout the GCN iterations. Based on the results, the following conclusions can be drawn:

Firstly, removing most of the edges makes the overall performance worse, validating the rationality of leveraging the four types of alignments between <text, image> pairs of mention and image to utilize fine-grained alignment relations.

Secondly, The substitution of dynamic edges with static ones causes a drastic performance drop on both datasets. This means that dynamic relation interaction has advantages over static alignments, which corroborates our motivation to employ dynamic GCN to model diverse alignments.

Thirdly, text-to-text edges have the greatest impact on performance compared to other edges. This is because we primarily rely on text for entity linking, while visual clues serve mainly as auxiliary information. As a result, text-to-text alignment contributes more to our model, which is consistent with our motivation.

Fourthly, compared to WikiDiverse, WikiMEL is less affected after removing a type of edges, especially for those linked to entity image vertices, i.e., image-to-image and text-to-image edges. This could be attributed to the fact that most entity images in WikiMEL are portraits of individuals, so they cannot provide substantial information for the MEL task, as our image encoder ResNet, which is trained on ImageNet, is unable to differentiate between various faces. Therefore, the exclusion of edges connected to entity images has a limited impact on the aggregation of crucial information.

\modify{Lastly, while removing edges causes a drastic performance drop on most metrics, we do observe a slight increase on the Top-20 metric in WikiMEL. This may be because the ranked entities in Top-20 metric have a bigger risk of noise. Compared to Top-1, Top-5, and Top-10, lower-ranked entities in Top-20 have lower relevance and are more likely to be noisy. In this case, removing edges will prevent noise propagation. Overall, our method outperforms other methods such as GHMFC on the Top-20 metric in WikiMEL, which indicates our proposed four fine-grained alignments bring more improvements over noise interference.}

\subsection{Effect of Hyper-parameter $L$}
We tune the value of hyper-parameters $L$ on the validation set of each dataset, and then evaluate the performance of the model on the test set. Table \ref{tab:hyper} shows the results when the number of GCN layers was separately set to 1, 2, 3, 4, and 5.

As the value of $L$ increased, the performance of DRIN improves, with the best results achieved when $L=2$. However, once the value of $L$ exceeds 2, performance does not continue to increase and even begin to decline. This is a common phenomenon in GCN. Firstly, a deep GCN is very difficult to train \cite{DBLP:conf/uai/Abu-El-HaijaKPL19}. Secondly, as GCN aggregates vertex information from neighbors, their representations tend to converge when the number of layers is too large \cite{DBLP:conf/aaai/LiHW18}, making it difficult for the model to distinguish between candidate entities.


\subsection{Case Study}
To better understand the advantages of our proposed method, we present qualitative results of DRIN compared with previous fusion-based methods and DRIN with static edges in a case study.

\begin{figure}
    \centering
    \includegraphics[width=0.4\textwidth]{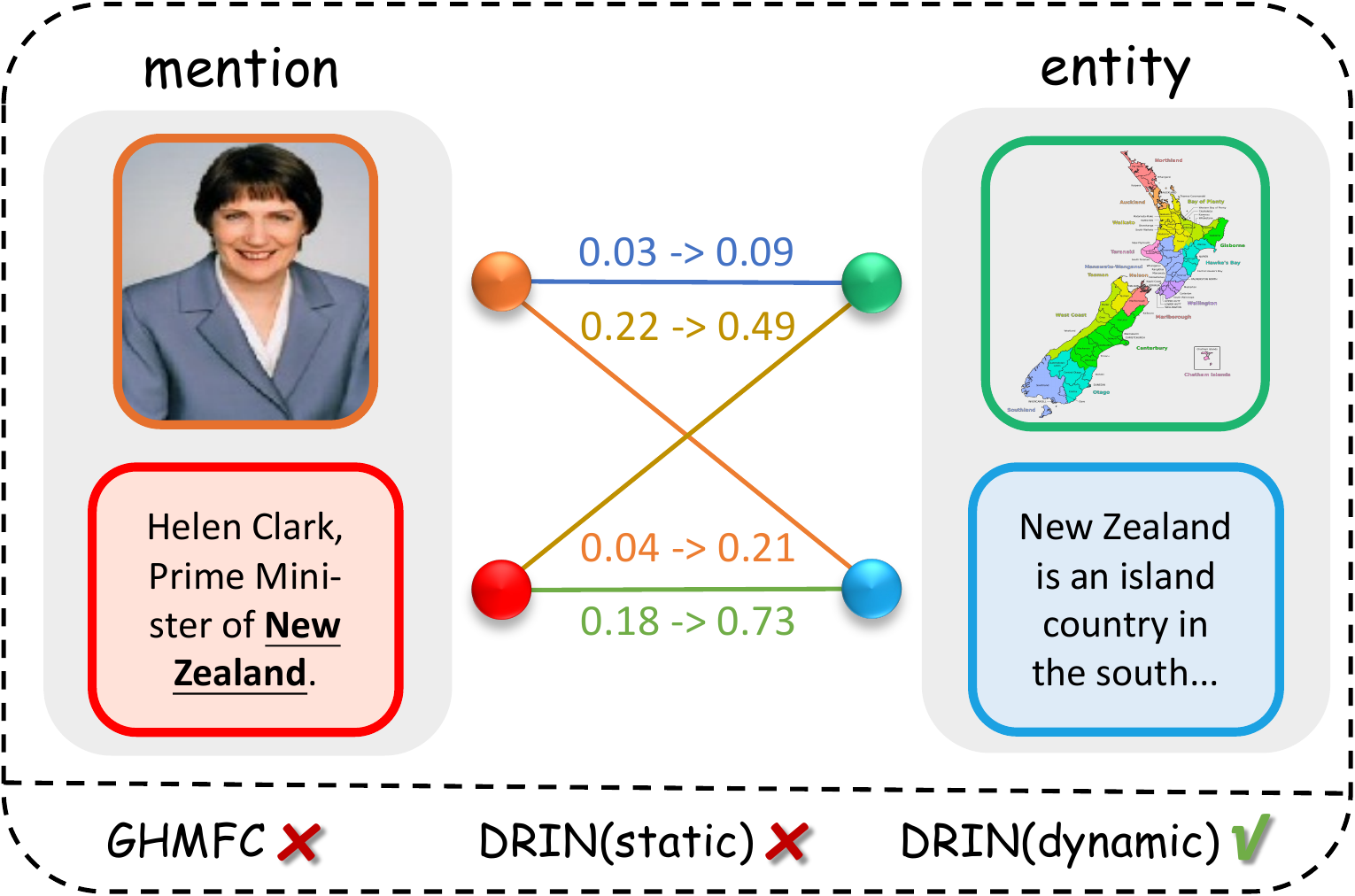}
    \caption{predictions of GHMFC and DRIN on a sample. This image contains the mention with its correct entity. The values before and after the arrow is the corresponding edge weights before and after being updated by GCN, respectively. \xmark, \cmark denote incorrect and correct predictions.}
    \label{fig:case}
\end{figure}

As shown in Figure \ref{fig:case}, the image of the mention (a photo of the prime minister) and its correct entity (a map of New Zealand) do not match. As a result, GHMFC which fuses the features of text and images tends to lower the matching probability of them, resulting in an incorrect prediction. Additionally, due to the low similarity between the text and images of mention and entity, the initial edge weights of DRIN are all small. Therefore, static DRIN cannot effectively aggregate information. However, DRIN, which dynamically models fine-grained alignments, understands that in this sample text-to-text and text-to-image alignments are more important, so the corresponding edges are enhanced and thus a more robust representation is obtained. Consequently, DRIN predicts a higher probability and solves this sample correctly.

\subsection{Complexity}
The time complexity of DRIN is $O\left(A(n^2d+nd^2)+12Lrd^2\right)$, where $n$ is the sequence length, $d$ is the dimension of the hidden state, and $A$ is a constant. This is because the multimodal feature extraction counts for $O(A(n^2d+nd^2))$; through a single GCN iteration, vertices and edges are updated with its neighbors, which counts for $O(d^2(2\times 2r+2r\times 2))$ and $O(d^2 \times 4r)$, respetively.

In comparison, the complexity of the previous fusion-based methods like GHMFC is $O((A+B)(n^2d+nd^2))$, where $B$ is a constant. Since $r,L$ are constants that is not very large (usually $A,B,n\sim 100;~r\le 100;~L\le 5$), the complexity of our method does not drastically exceed that of previous works in the order of magnitude.

\section{Conclusion}
In this paper, we propose a novel Dynamic Relation Interactive Network (DRIN) for the Multimodal Entity Linking (MEL) task. The main idea of our approach is to explicitly and dynamically model four kinds of fine-grained alignments between mention and entity to enhance their representation. Results from experiments indicate that our model achieves far better performance than other state-of-the-art methods.

\begin{acks}
We would like to thank the anonymous reviewers for their constructive comments. This work was supported by the National Natural Science Foundation of China (No. 62206126 and No. 61976114).
\end{acks}



\bibliographystyle{ACM-Reference-Format}
\bibliography{sample-base-new}

\end{document}